# Efficient Test Selection in Active Diagnosis via Entropy Approximation

**Alice X. Zheng**
Department of EECS
U. C. Berkeley
Berkeley, CA 94720-1776
alicez@eecs.berkeley.edu

**Irina Rish**
IBM T.J. Watson Research Center
19 Skyline Drive
Hawthorne, NY 10532
rish@us.ibm.com

**Alina Beygelzimer**
IBM T.J. Watson Research Center
19 Skyline Drive
Hawthorne, NY 10532
beygel@us.ibm.com

## Abstract

We consider the problem of diagnosing faults in a system represented by a Bayesian network, where diagnosis corresponds to recovering the most likely state of unobserved nodes given the outcomes of tests (observed nodes). Finding an optimal subset of tests in this setting is intractable in general. We show that it is difficult even to compute the next most-informative test using greedy test selection, as it involves several entropy terms whose exact computation is intractable. We propose an approximate approach that utilizes the loopy belief propagation infrastructure to simultaneously compute approximations of marginal and conditional entropies on multiple subsets of nodes. We apply our method to fault diagnosis in computer networks, and show the algorithm to be very effective on realistic Internet-like topologies. We also provide theoretical justification for the greedy test selection approach, along with some performance guarantees.

## 1 Introduction

The problem of fault diagnosis appears in many places under various guises. Examples include medical diagnosis, computer system troubleshooting, decoding messages sent through a noisy channel, etc. In recent years, diagnosis has often been formulated as an inference problem on a Bayesian network, with the goal of assigning most likely states to unobserved nodes based on outcome of test nodes.

An important issue in diagnosis is the trade-off between the cost of performing tests and the achieved accuracy of diagnosis. It is often too expensive or even impossible to perform all tests. In this paper, we concentrate on the problem of *active* diagnosis, in which tests are selected sequentially to minimize the cost of testing. We use entropy as the cost function and select a set of tests providing maximum information, or minimum conditional entropy, about the unknown variables.

However, exact computation of conditional entropies in a general Bayesian network can be intractable. While much existing research has addressed the problem of efficient and accurate probabilistic inference, other probabilistic quantities, such as conditional entropy and information gain, have not received nearly as much attention. There is a vast amount of literature on value-of-information and most-informative test selection [10, 4, 9, 11], but none of the previous work appears to focus on the computational complexity of most-informative test selection in a general Bayesian network setting.

We propose an approximation algorithm for computing marginal conditional entropy. The algorithm is based on loopy belief propagation, a successful approximate inference method. We illustrate the algorithm at work in the setting of fault diagnosis for distributed computer networks, and demonstrate promising empirical results. We also apply existing theoretical results on the optimality of certain greedy algorithms to our test selection problem, and analyze the effect of approximation error on the expected cost of active diagnosis. Our method is general enough to apply to other applications of Bayesian networks that require the computation of information gain and conditional entropies of subsets of nodes. In our application, it can efficiently compute the information gain for all candidate tests simultaneously.

The paper is structured as follows. Section 2 introduces necessary background and definitions. In section 3, we describe the general problem of active diagnosis and the computational complexity issue thereof. We propose a solution to this problem in section 4. Section 5 discusses an application of our approach in the context of distributed computer system diagnosis, while section 6 presents empirical results. We survey related work in section 7, and conclude in section 8.

## 2 Background and Definitions

Let $\mathbf{X} = \{X_1, X_2, \ldots, X_N\}$ denote a set of $N$ discrete random variables and $\mathbf{x}$ a possible realization of $\mathbf{X}$. A *Bayesian network* is a directed acyclic graph (DAG) $G$ with nodes corresponding to $X_1, X_2, \ldots, X_N$ and edges representing direct dependencies [16]. The dependencies are quantified by associating each node $X_i$ with a local conditional probability distribution $P(x_i \mid \mathbf{pa}_i)$, where $\mathbf{pa}_i$ is an assignment to the parents of $X_i$ (nodes pointing to $X_i$ in the Bayesian network). The set of nodes $\{x_i, \mathbf{pa}_i\}$ is called a *family*. The joint probability distribution function (PDF) over $\mathbf{X}$ is given as product

$$P(\mathbf{x}) = \prod_{i=1}^{N} P(x_i \mid \mathbf{pa}_i). \quad (1)$$

We use $\mathbf{E} \subseteq \mathbf{X}$ to denote a possibly empty set of *evidence* nodes for which observation is available.

For ease of presentation, we will also use the terminology of *factor graphs* [6], which unifies directed and undirected graphical representations of joint PDFs. A factor graph is an undirected bipartite graph that contains factor nodes (usually shown as squares) and variable nodes (shown as circles). (See Fig. 1 for an example.) There is an edge between a variable node and a factor node if and only if the variable participates in the *potential function* of the corresponding factor. The joint distribution is assumed to be written in a factored form

$$P(\mathbf{x}) = \frac{1}{Z} \prod_{a} f_a(\mathbf{x}_a), \quad (2)$$

where $Z$ is a normalization constant called the *partition function*, and the index $a$ ranges over all factors $f_a(\mathbf{x}_a)$, defined on the corresponding subsets $\mathbf{X}_a$ of $\mathbf{X}$.

The computation complexity of many probabilistic inference problems can be related to graphical properties. Exact inference algorithms require time and space exponential in the *treewidth* [16] of the graph, which is defined to be the size of the largest clique induced by inference, and can be as large as the size of the graph. Many common probabilistic inference problems are NP-complete. [1] This includes our problem of probabilistic diagnosis, which can be formulated as a *Maximum A Posteriori* (MAP) probability problem: given a set of observations, find the most likely states of unobserved variables.

Although probabilistic inference can be intractable in general, there exists a simple linear-time approximate inference algorithm known as *belief propagation (BP)* [16]. BP is provably correct on polytrees (i.e. Bayesian networks with no undirected cycles), and can be used as an approximation on general networks. In belief propagation, probabilistic messages are iterated between the nodes. The process could diverge; convergence is guaranteed only for polytrees.

Let $a$ denote a factor node and $i$ one of its variable nodes. Let $\mathcal{N}(a)$ represent the neighbors of $a$, i.e., the set of variable nodes connected to that factor. Let $\mathcal{N}(i)$ denote the neighbors of $i$, i.e., the set of factor nodes to which variable node $i$ belongs. The BP message from node $i$ to factor $a$ is defined as (see, e.g., [12])

$$n_{i \to a}(x_i) := \prod_{c \in \mathcal{N}(i) \backslash a} m_{c \to i}(x_i), \quad (3)$$

and the message from factor $a$ to node $i$ is defined as

$$m_{a \to i}(x_i) := \sum_{\mathbf{x}_a \backslash x_i} f_a(\mathbf{x}_a) \prod_{j \in \mathcal{N}(a) \backslash i} n_{j \to a}(x_j). \quad (4)$$

Based on these messages, we can compute the beliefs for each node and the probability potential for each factor:

$$b_i(x_i) \propto \prod_{a \in \mathcal{N}(i)} m_{a \to i}(x_i), \quad (5)$$

$$b_a(\mathbf{x}_a) \propto f_a(\mathbf{x}_a) \prod_{i \in \mathcal{N}(a)} n_{i \to a}(x_i). \quad (6)$$

Observations are incorporated into the process via $\delta$-functions as local potentials for the evidence nodes. In that case, $b_i(x_i)$ becomes the approximation of the posterior probability $P(x_i \mid \mathbf{e})$.

## 3 The Active Test Selection Problem

In many diagnosis problems, the user has an opportunity to actively select tests in order to improve the accuracy of diagnosis. For example, in medical diagnosis, doctors face the *experiment design* problem of choosing which medical tests to perform next.

Let $\mathbf{S} = \{S_1, S_2, \ldots, S_N\}$ denote a set of unobserved random variables we wish to diagnose, and let $\mathbf{T} = \{T_1, T_2, \ldots, T_M\}$ denote the available set of tests. Our objective is to maximize diagnostic quality while minimizing the cost of testing. The diagnostic quality of a subset of tests $\mathbf{T}^*$ can be measured by the amount of uncertainty about $\mathbf{S}$ that remains after observing $\mathbf{T}^*$. From the information-theoretic perspective, a natural measurement of uncertainty is the conditional entropy $H(\mathbf{S} \mid \mathbf{T}^*)$. Clearly, $H(\mathbf{S} \mid \mathbf{T}) \leq H(\mathbf{S} \mid \mathbf{T}^*)$ for all $\mathbf{T}^* \subseteq \mathbf{T}$. Thus the problem is to find $\mathbf{T}^* \subseteq \mathbf{T}$ which minimizes both $H(\mathbf{S} \mid \mathbf{T}^*)$ and the cost of testing. When all tests have equal cost, this is equivalent to minimizing the number of tests.

This problem is known to be NP-hard [19]. A simple greedy approximation is to choose the next test to be $T^* = \arg\min_T H(\mathbf{S} \mid T, \mathbf{T}')$, where $\mathbf{T}'$ is the currently selected test set. The expected number of tests produced by the greedy strategy is known to be within a $O(\log N)$ factor from optimal (see Appendix). The same result holds for approximations (within a constant multiplicative factor) to the greedy approach. Furthermore, our empirical results show that the approach works well in practice.

We make a distinction between off-line test selection and online test selection. In online selection, previous test outcomes are available when selecting the next test. Off-line test selection attempts to plan a suite of tests before any observations have been made. We will focus on the online approach, sometimes called *active diagnosis*, which is typically much more efficient in practice than its off-line counterpart [19].

**Active Test Selection Problem:** Given the observed outcome $\mathbf{t}'$ of previously selected sequence of tests $\mathbf{T}'$, select the next test to be $\arg\min_T H(\mathbf{S} \mid T, \mathbf{t}')$.

In a Bayesian network, the joint entropy $H(\mathbf{X})$ can be decomposed into sum of entropies over the families and thus can be easily computed using the input potential functions. Conditional marginal entropies, on the other hand, do not generally have this property. Under certain independence conditions they decompose into functions over the families. But computing those functions will require inference. (See Appendix for proofs.)

**Lemma 1.** *Given a Bayesian network representing a joint PDF $P(\mathbf{X})$, the joint entropy $H(\mathbf{X})$ can be decomposed into the sum of entropies over the families: $H(\mathbf{X}) = \sum_{i=1}^{N} H(X_i \mid \mathbf{Pa_i})$.* □

**Lemma 2.** *Given a Bayesian network representing a joint PDF $P(\boldsymbol{S}, \boldsymbol{T})$, where $\forall i : \mathbf{pa}_{T_i} \subseteq \boldsymbol{S}$ (i.e. tests $T_i$ and $T_j$ are independent given a subset of **S**), the observation $\boldsymbol{t}'$ of previously selected test set, and a candidate test $T$, the conditional marginal entropy $H(\boldsymbol{S} \mid T, \boldsymbol{t}')$ can be written as*

$$H(\boldsymbol{S} \mid T, \boldsymbol{t}') = -\sum_{t, \boldsymbol{s}_{\mathbf{pa}_T}} P(\boldsymbol{s}_{\mathbf{pa}_T}, t \mid \boldsymbol{t}') \log P(t \mid \boldsymbol{s}_{\mathbf{pa}_T}) + \sum_{t} P(t \mid \boldsymbol{t}') \log P(t \mid \boldsymbol{t}') + const, \quad (7)$$

*where* const *is a constant expression.* □

Minimizing conditional entropy is a particular instance of *value-of-information* (VOI) analysis [9], where tests are selected to minimize the expected value of a certain *cost function* $c(\mathbf{s}, t, \mathbf{t}')$. The result of Lemma 2 can be generalized to this case if the cost function is decomposable over the families. See Lemma 3 in the Appendix for details.

Since observations of test outcome correlate the parent nodes, the exact computation of all the posterior probabilities in Eqn. (7) is intractable. We can certainly use an existing approximation method to compute $P(\mathbf{s}_{\mathbf{pa}_T}, t \mid \mathbf{t}')$ and $P(t \mid \mathbf{t}')$. But a more efficient approach is possible if we exploit the belief propagation infrastructure.

## 4 BP for Entropy Approximation

Let us consider the problem of computing the conditional marginal entropy

$$H(\mathbf{X}_a \mid \mathbf{e}) = -\sum_{\mathbf{x}_a} P(\mathbf{x}_a \mid \mathbf{e}) \log P(\mathbf{x}_a \mid \mathbf{e}), \quad (8)$$

where $P(\mathbf{x}_a \mid \mathbf{e}) = \sum_{\mathbf{x} \backslash \mathbf{x}_a} P(\mathbf{x} \mid \mathbf{e})$, $\mathbf{x} \backslash \mathbf{x}_a$ representing variable nodes not in $\mathbf{x}_a$. The trick is to replace the marginal posterior $P(\mathbf{x}_a \mid \mathbf{e})$ with its factorized BP approximation, and make use of the BP message passing mechanism to perform the summation over $\mathbf{x}_a$. We call this process Belief Propagation for Entropy Approximation (BPEA).

Pick any node $X_0$ from $\mathbf{X}_a$ and designate it as the root node. We modify the final message passed to $X_0$ as follows:

$$m'_{a \to 0}(x_0) := -\sum_{\mathbf{x}_a \backslash x_0} \tilde{b}_a(\mathbf{x}_a) \log \tilde{b}_a(\mathbf{x}_a). \quad (9)$$

Here, $\tilde{b}_a(\mathbf{x}_a)$ is the unnormalized belief of $X_a$ (i.e., $\tilde{b}_a(\mathbf{x}_a) = \sigma b_a(\mathbf{x}_a)$, where $\sigma = \sum_{\mathbf{x}_a} \tilde{b}_a(\mathbf{x}_a)$).

Plugging in $\tilde{b}_a(\mathbf{x}_a)$ in place of $P(\mathbf{x}_a \mid \mathbf{e})$ in Eqn. 8, we see that it only remains to sum over the root node $X_0$ and normalize properly.

$$\tilde{h}(\mathbf{X}_a \mid \mathbf{e}) := \sum_{x_0} m'_{a \to 0}(x_0), \quad (10)$$

$$h(\mathbf{X}_a \mid \mathbf{e}) := \frac{\tilde{h}(\mathbf{X}_a \mid \mathbf{e})}{\sigma} + \log \sigma. \quad (11)$$

It follows immediately that BPEA is exact whenever BP is exact.

The normalization constant $\sigma$ is already computed during normal BP iterations. The computation of $\tilde{b}_a(\cdot)$, $m'_{a \to i}$, and $\tilde{h}(\cdot)$ can all be piggy-backed onto the same BP infrastructure, and therefore does not impact its overall complexity. Furthermore, due to the local and parallel message update procedure in BP, we can compute the marginal posterior entropies of multiple families in one single sweep. This is an important advantage for the active probing setup.

It is also easy to show that the approach is extendible beyond the entropy computation, to an arbitrary cost function decomposable over families (see Lemma 3 in the Appendix). The cost function replaces the negative logarithm in Eqns. (8) and (9).

## 5 Application: Fault Diagnosis in Computer Networks

Suppose we wish to monitor a system of networked computers. Let $\mathbf{S}$ represent the binary state of $N$ network elements. $S_i = 0$ indicates that the element is in normal operation mode, and $S_i = 1$ indicates that the element is faulty. We can take $S_i$ to be any system component whose state can be measured using a suite of tests. If the system is large, it is often impossible to test each individual component directly. A common solution is to test a subset of components with a single *test probe*. If all the test components are okay, the test would return a 0. Otherwise the test would return 1, but it does not reveal which components are faulty.

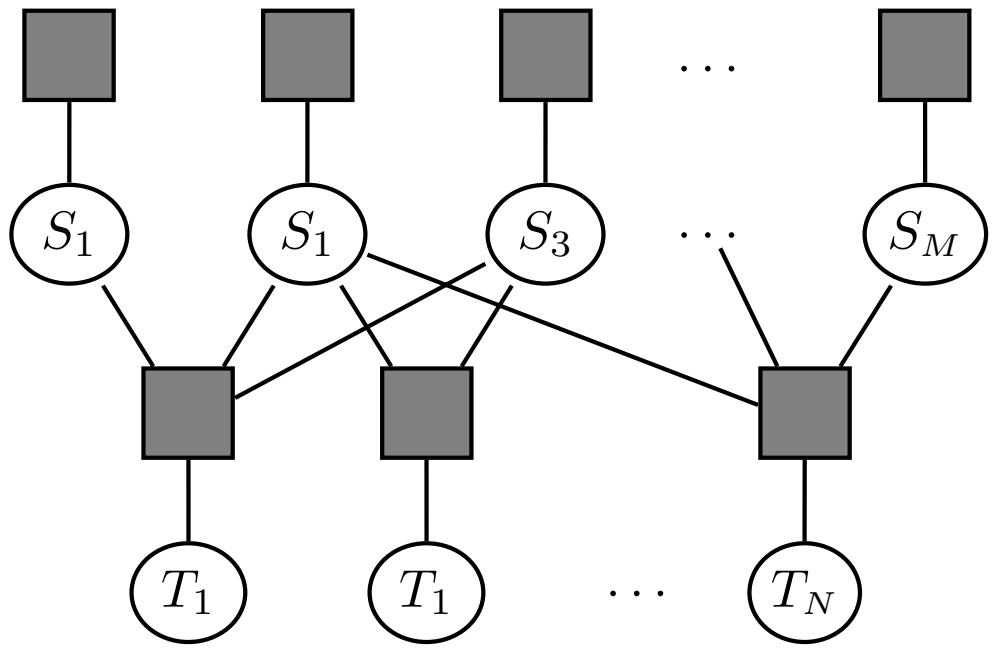

Figure 1: Factor graph of the fault diagnostic Bayes net.

We assume there are machines designated as *probe stations*, which are instrumented to send out *probes* to test the response of the network elements represented by **S**. Let **T** denote the available set of probes. A probe can be as simple as a *ping* request, which detects network availability. A more sophisticated probe might be an e-mail message or a webpage-access request. In the absence of noise a probe is a disjunctive test: it fails if an only if there is at least one failed node on its path. More generally, it is a noisy-OR test [16]. The joint PDF of all tests and network nodes forms the well-known QMR-DT model [13]:

$$P(s_j) = (\alpha_j)^{s_j}(1-\alpha_j)^{(1-s_j)}, \tag{12}$$

$$P(t_i = 0 \mid \mathbf{s}_{\mathbf{pa}_i}) = \rho_{i0} \prod_{j \in \mathbf{pa}_i} \rho_{ij}^{s_j}, \tag{13}$$

$$P(\mathbf{s}, \mathbf{t}) = \prod_i P(t_i \mid \mathbf{s}_{\mathbf{pa}_i}) \prod_j P(s_j). \tag{14}$$

Here, $\alpha_j := P(s_j = 1)$ is the prior fault probability, $\rho_{ij}$ is the so-called inhibition probability, and $(1-\rho_{i0})$ is the leak probability of an omitted faulty element. The inhibition probability is a measurement of the amount of noise in the network. Fig. 1 shows a factor graph representation of our model.

As discussed in Section 3, we adopt the *active probing* framework for fault diagnosis, sequentially selecting probes to minimize the conditional entropy. Our previous work [17] makes the single-fault assumption, which effectively reduces **S** to one random variable with $N+1$ possible states. In general, however, multiple faults could exist in the system simultaneously, which requires the more complicated conditional entropy given in Eqn. (7).

Let $A(T, \mathbf{S}_{\mathbf{pa}_T} \mid \mathbf{t}')$ denote the first term in Eqn. (7). This is the cross entropy between the posterior probability of $T$ and its parents, and the conditional probability of $T$ given its parents. The second term in Eqn. (7) is simply the negative conditional entropy $-H(T \mid \mathbf{t}')$.

We deal with the two entropy terms separately. For $H(T \mid \mathbf{t}')$, we may use approximation methods such as BP or GBP to calculate the belief $b(t \mid \mathbf{t}')$, which can then be used to directly compute $H(T \mid \mathbf{t}')$. (Note that the summation over values of $T$ is simple since $T$ is binary-valued.) To calculate $A(T, \mathbf{S}_{\mathbf{pa}_T} \mid \mathbf{t}')$, we use the entropy approximation method BPEA, as described in Section 4. Because BP message updates are done locally, we can compute $A(T, \mathbf{S}_{\mathbf{pa}_T} \mid \mathbf{t}')$ for all unobserved $T$ nodes during a single application of BP. Thus, picking the next probe requires only one run of the BPEA approximation algorithm.

For each candidate probe, we designate the probe node $T$ itself as the root node. The unnormalized belief has the form

$$\tilde{b}_t(t, \mathbf{s}_{\mathbf{pa}_T}) := P(t \mid \mathbf{s}_{\mathbf{pa}_T}) \prod_{j \in \mathbf{pa}_T} n_{j \to t}(s_j). \tag{15}$$

This is used to calculate the modified message $m'_{a \to t}(t)$ (cf. Eqn. (9)). However, since $A(T, \mathbf{S}_{\mathbf{pa}_T} \mid \mathbf{t}')$ is a cross entropy term, we do not take the $\log$ of $\tilde{b}$, but rather take the logarithm of the known probabilities $P(t \mid \mathbf{s}_{\mathbf{pa}_T})$. This simplifies the normalization step described in Eqn. (11) to $A(T, \mathbf{S}_{\mathbf{pa}_T} \mid \mathbf{t}') = \tilde{A}(T, \mathbf{S}_{\mathbf{pa}_T} \mid \mathbf{t}')/\sigma$, where $\sigma = \sum_{t, \mathbf{s}_{pa}(T)} \tilde{b}_t(t, \mathbf{s}_{\mathbf{pa}_T})$.

## 6 Empirical Results

We conduct experiments on network topologies built by the INET generator [20], which simulates an Internet-like topology at the Autonomous Systems level. Our dataset includes a set of networks of 485 nodes, where the number of probe stations varies from 1 to 50.

The connections between probe nodes and network nodes are generated with two goals in mind: detection and diagnosis. A detection probe set needs to cover all network components, so that at least one probe has a positive probability of returning 1 when a component fails. A diagnosis probe set needs to further distinguish between faulty components. Optimal probe set design is NP-hard for either detection or diagnosis. For the datasets used here, we first use a greedy approach to obtain a probe set that covers all network components, then augment this set with additional probes in order to guarantee single-fault diagnosis. Interested readers may find detailed discussions of probe set design for diagnostic Bayesian networks in [11, 18].

In our experiments, we measure the effects of prior fault probability $\alpha$ and inhibition probability $\rho$ on approximation and diagnostic quality. We compare the approximate entropy values and the quality of the selected probe set against the ground truth, which is obtained via the junction tree exact inference algorithm. In subsection 6.3, we also summarize how the type of network may effect computational efficiency. Since all measurements depend on the particular set of probe outcomes, we repeat all experiments on 10 different samples of the Bayes net.

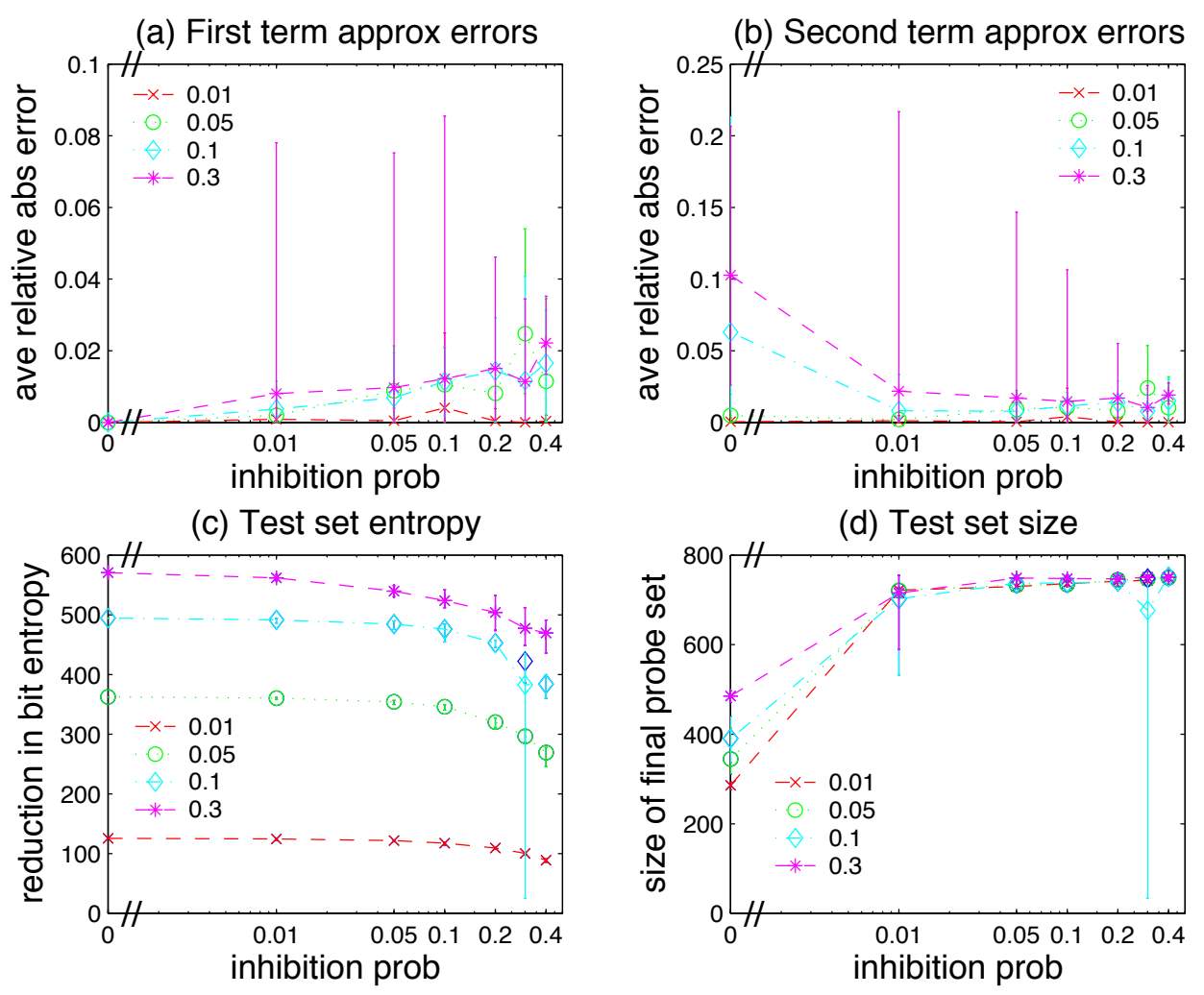

Figure 2: Approximation errors and diagnostic quality for an augmented detection network. Each curve represents a different prior fault probability.

We use the diagnostic quality of the probe set to determine when to stop the probe selection process: when the reduction in entropy for the past 5 iterations is no more than 0.00001, the selection process is deemed to converge. Otherwise we continue until all probes have been picked.

## 6.1 Approximation accuracy

First, we look at approximation accuracy. Recall that at each time step of the active probing process, we obtain a vector of approximate entropy values, one for each candidate probe $T$. We average the relative error between the approximate values and the exact values for all candidate probes, and further average over all time steps and samples. Let $M$ denote the total number of probes, $n$ the number of selected probes, $h_{ij}$ the approximate value for probe $j$ at the $i$th time step of probe selection, and $H_{ij}$ the corresponding exact value. We compute

$$R(h,H) := \frac{1}{n}\sum_{i=0}^{n-1}\frac{1}{M-i}\sum_{j=1}^{M-i}\frac{|h_{ij}-H_{ij}|}{|H_{ij}|}. \qquad (16)$$

We conduct this experiment on the detection network with 10 probe stations, augmented with single-node probes. Fig. 2(a-b) contains plots of the average, the minimum, and the maximum approximation errors, taken over 10 samples of probe outcomes. Relative error values are shown separately for the first term, $A(T, \mathbf{S}_{\mathbf{pa}_T} \mid \mathbf{t}')$, and the second term, $H(T \mid \mathbf{t}')$. For both terms, the approximation errors are generally lower at lower $\alpha$ values. The average errors do not exceed 2%, with the only exception being the BP error for term two at $\alpha = 0.3$ and $\rho = 0$, which reaches up to 10%. BP approximaton errors of the second term seem to be generally higher than BPEA approximations of the first term. At the maximum, the approximation error never exceeds 10% for term one, and 20% for term two. BP errors for term two does not seem to contain any linear trends with respect to $\rho$. However, BPEA's approximation quality of term one does seem to become slightly worse at higher levels of the inhibition probability.

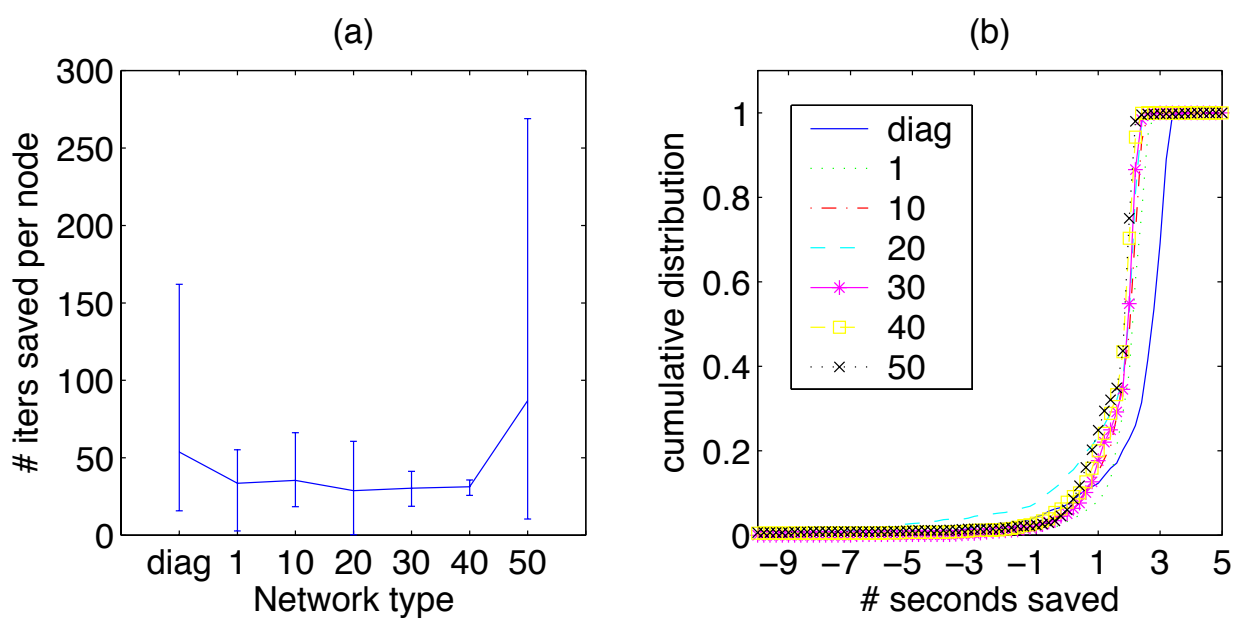

Figure 3: Efficiency of approximate method. (a) Average number of BP iterations saved by re-using messages; (b) CDF of speed-up (in CPU seconds) compared to exact method.

## 6.2 Diagnostic quality

The quality of diagnosis is taken to be the reduction in conditional bit entropy of the hidden states. If $\mathbf{t}'$ represents the observed outcomes of the final set of selected probes, we measure $H(\mathbf{S}) - H(\mathbf{S} \mid \mathbf{t}') = -\sum_{\mathbf{s}} P(\mathbf{s}) \log_2 P(\mathbf{s}) + \sum_{\mathbf{s}} P(\mathbf{s} \mid \mathbf{t}') \log_2 P(\mathbf{s} \mid \mathbf{t}')$.

Fig. 2(c) compares the diagnostic quality of approximate and exact algorithms on the augmented detection network with 10 probe stations. Overall, the reduction in bit entropy is larger for higher values of $\alpha$. This is due to the fact that $H(\mathbf{S})$ is higher when $\alpha$ is larger. The quality of the exact algorithm is almost identical to that of the approximate algorithm. The two are virtually indistinguishable, except at $\alpha = 0.1$ and $\rho = 0.3$. There is an outlier at this combination. For one of the samples, the value of the entropy $H(\mathbf{S} \mid \mathbf{t}')$ plateaued unusually early during the active probing process, fooling the algorithm into believing that it had converged, even though the amount of *reduction* in entropy is still very small. Fig. 2(d) shows that the process terminated after selecting only a small set of tests. This outlier is an artifact of our convergence criterion, not of the approximate algorithm itself.

Fig. 2(d) looks at the size of the final selected probe set when active probing converges. Here again, the two algorithms have almost identical behavior. The value of $\alpha$ does not have much impact on the number of selected tests, except when $\rho = 0$ (i.e., no noise in the tests), in which case fewer tests are needed for diagnosis at lower levels of $\alpha$.

These results demonstrate that, while the approximated entropy values may deviate from the truth, the diagnostic quality of the approximate method is virtually identical to that obtained using the exact method. Combined with its speed advantages as described in the next section, these results make a strong case for why the approximate method is preferable over the exact one.

### 6.3 Implementation and speed

We use the junction tree inference engine in Kevin Murphy's Bayes Net Toolbox [15] for Matlab to obtain exact singleton posterior probabilities. The approximate method is implemented on top of the belief propagation C++/mex code developed by Yair Weiss and Talya Meltzer. Additionally, we speed up the approximate active probing process by re-using BP messages at the start of each round of test selection, thereby maintaining BP's state from the end of the selection round. We find that BP converges in substantially fewer iterations this way.

Fig. 3(a) plots the average, maximum, and minimum number of BP iterations that we save by re-using BP messages. The results are aggregated over 5 samples of the Bayes net. The x-axis denotes the type of network used. The label `diag` represents the diagnosis network with 1 probe station, and the rest are detection networks with various numbers of probe stations. In the detection network with 50 probe stations, we save up to 269 iterations per test node at the maximum. On average, re-using messages shortens the BP convergence time by 40-50 iterations *per test*. If active probing selected 100 tests, say, then re-using messages would require 4000 to 5000 fewer iterations of belief propagation.

Fig. 3(b) is a plot of the empirical cumulative distribution of the speed-up using the approximate method. For all of the detection networks, the approximate method is at least 1 CPU second faster than the exact method for $75\%$ of the test nodes. The speed-up is even higher for the diagnostic network, where for $78\%$ of all test nodes the approximate method saves at least 2 CPU seconds *per node*. This amounts to substantial savings over the entire active probing process. Also keep in mind that, for networks with large tree-width, the exact method is not even computationally feasible. Hence, approximation may be the only realistic option.

## 7 Related Work

The problem of most-informative test selection was previously addressed in various areas including diagnosis, decision analysis, and feature selection in machine learning. Given a cost function, a common decision-theoretic approach is to compute the *expected value-of-information* [10] of a candidate test, i.e., the expected cost of making a decision after observing the test outcome. When entropy is used as the cost function, the approach is called most-informative test selection. In particular, most-informative test selection was considered in the context of model-based diagnosis [4] and probabilistic diagnosis [16].

Previous research [9, 8] on VOI analysis has made various simplifying assumptions such as binary hypothesis and direct observations. An interesting but tangential approach was taken in [9], which proposes to select a set of tests based on a law-of-large-numbers approximation of the VOI. Up to now, however, no one seems to have addressed the efficiency of computing single-test information gain in a generic Bayesian network.

Most-informative test selection is quite similar to the optimal coding problem [2]. Namely, the hidden state vector **S** is the input message, and the test outcomes **T** the output message from some noisy channel. The goal of most-informative test selection is to minimize the number of bits sent through the channel while still accurately decoding the input message. There is, however, an important difference between the two. In the coding domain, one may separate source coding from channel coding. Fault diagnosis, on the other hand, has to deal with a combination of the two, represented by the conditional probability $P(T_i \mid \mathbf{S}_{\mathbf{pa}_i})$. We may have no control over the source coding function, but we can still aim to select the smallest, most informative subset of tests.

In the context of probing, optimal test selection is very similar to the *group testing* problem [5]. Given a set of Boolean variables, the objective of group testing is to find all 'failed' objects by using a sequence of disjunctive tests. Particularly, sequential test selection is known as *adaptive group testing* [5]. (There is also a direct connection between adaptive group testing and Golomb codes [7].) Note that group testing assumes no constraints on the tests (i.e., any subset of objects can be tested together), while in Bayesian networks the tests can be only selected from a fixed set. Even in a less restrictive case of probe selection, we are still constrained by the network topology. Theoretical analysis of constrained group testing is difficult.

## 8 Conclusions

We propose an entropy approximation method based on loopy belief propagation, and examine its behavior on the application of active probing for fault diagnosis in a networked computer system. The level of approximation error varies slightly with the level of noise. But even so, the diagnosis quality is practically identical to that of the exact method. Furthermore, the approximate method can handle larger networks than the exact method, and is almost always faster on the smaller ones. This highlights a promising direction for active probing and fault diagnosis, as well as for entropy approximation on Bayesian networks in general.

## Appendix

**Lemma 1.** *Proof:* From Eqn. (1) we get

$$H(\mathbf{X}) = -\sum_{\mathbf{x}} P(\mathbf{x}) \sum_{i=1}^{N} \log P(x_i \mid \mathbf{pa}_i)$$

$$= -\sum_{i=1}^{N} \sum_{x_i, \mathbf{pa}_i} P(x_i, \mathbf{pa}_i) \log P(x_i \mid \mathbf{pa}_i) = \sum_{i=1}^{N} H(X_i \mid \mathbf{Pa_i}).$$

□

**Lemma 2.** *Proof:* $H(\mathbf{S} \mid T, \mathbf{t}') = H(\mathbf{S}, T \mid \mathbf{t}') - H(T \mid \mathbf{t}')$ where $H(T \mid \mathbf{t}') = -\sum_t P(t \mid \mathbf{t}') \log P(t \mid \mathbf{t}')$, and

$$H(\mathbf{S}, T \mid \mathbf{t}') = -\sum_{\mathbf{s},t} P(\mathbf{s}, t \mid \mathbf{t}') \log P(\mathbf{s}, t \mid \mathbf{t}')$$

$$= -\sum_{\mathbf{s},t} P(\mathbf{s}, t \mid \mathbf{t}') \log P(\mathbf{s}, t, \mathbf{t}') + \sum_{\mathbf{s},t} P(\mathbf{s}, t \mid \mathbf{t}') \log P(\mathbf{t}')$$

$$= -\sum_{\mathbf{s},t} P(\mathbf{s}, t \mid \mathbf{t}') \log P(\mathbf{s}, t, \mathbf{t}') + \log P(\mathbf{t}'). \quad (17)$$

The last term in Eqn. (17) is independent of $T$ and can be replaced by $const$. Since the $T$ nodes are conditionally independent given their parents, $P(\mathbf{s}, t, \mathbf{t}') = P(t \mid \mathbf{s}_{\mathbf{pa}_T}) \prod_j P(t'_j \mid \mathbf{s}_{\mathbf{pa}_j}) P(\mathbf{s})$. This assumption is essential to the proof, because in general $P(\mathbf{s}, t, \mathbf{t}')$ may not factorize, and no further simplifications would be possible.

The first term in Eqn. (17) can now be written as $\sum_{\mathbf{s},t} P(\mathbf{s}, t \mid \mathbf{t}') \log P(t \mid \mathbf{s}_{\mathbf{pa}_T}) + \sum_{\mathbf{s}} P(\mathbf{s} \mid \mathbf{t}') \log P(\mathbf{s}) \prod_j P(t'_j \mid \mathbf{s}_{\mathbf{pa}_j})$, where the second term does not involve $T$, and the first term can be simplified as $\sum_{\mathbf{s}_{\mathbf{pa}_T}, t} P(\mathbf{s}_{\mathbf{pa}_T}, t \mid \mathbf{t}') \log P(t \mid \mathbf{s}_{\mathbf{pa}_T})$. Finally,

$$H(\mathbf{S} \mid T, \mathbf{t}') = -\sum_{t, \mathbf{s}_{\mathbf{pa}_T}} P(\mathbf{s}_{\mathbf{pa}_T}, t \mid \mathbf{t}') \log P(t \mid \mathbf{s}_{\mathbf{pa}_T}) +$$

$$\sum_t P(t \mid \mathbf{t}') \log P(t \mid \mathbf{t}') + \text{ const.} \quad \square$$

**Lemma 3.** *Given a Bayesian network representing a joint PDF $P(\boldsymbol{S}, \boldsymbol{T})$, where $\forall i, j$, $T_i$ and $T_j$ are independent given some subset of $\boldsymbol{S}$, and a cost function where $c(t, \boldsymbol{s} \mid \boldsymbol{t}') = c(t, \boldsymbol{s}_{\mathbf{pa}_T})$, the expected cost of choosing test $t$ can be written as*

$$E_{P(\boldsymbol{s},t|\boldsymbol{t}')} c(t, \boldsymbol{s} \mid \boldsymbol{t}') = \sum_{t, \boldsymbol{s}_{\mathbf{pa}_T}} P(\boldsymbol{s}_{\mathbf{pa}_T}, t \mid \boldsymbol{t}') c(t, \boldsymbol{s}_{\mathbf{pa}_T}), \quad (18)$$

*where $\boldsymbol{t}'$ is an observation of previously selected tests, and $T$ a candidate test.*

*Proof:* $E_{P(\mathbf{s},t|\mathbf{t}')} c(t, \mathbf{s} \mid \mathbf{t}') = \sum_{\mathbf{s},t} P(\mathbf{s}, t \mid \mathbf{t}') c(t, \mathbf{s} \mid \mathbf{t}') = \sum_{\mathbf{s},t} P(\mathbf{s}, t \mid \mathbf{t}') c(t, \mathbf{s}_{\mathbf{pa}_T}) = \sum_{\mathbf{s}_{\mathbf{pa}_T}, t} P(\mathbf{s}_{\mathbf{pa}_T}, t \mid \mathbf{t}') c(t, \mathbf{s}_{\mathbf{pa}_T})$. □

### Approximation Quality of the Greedy Strategy

Recall that we have a set $\{S_1, \ldots, S_N\}$ of binary variables representing the state of $N$ elements, and the goal is to decode this state using tests.

For simplicity, we will only consider the case when tests correspond to deterministic disjunctions, i.e., all inhibition probabilities and the leak probability are zero. Thus an outcome of a test splits the state space into states that are consistent with the outcome and those that are not.

If any combination of the elements can be faulty, the system can, in principle, be in any one of $2^N$ possible states. However, the effective state space of a test $T$ involving $n$ elements contains $2^n$ "states," each corresponding to $2^{N-n}$ states in the original state space $\{0,1\}^N$. If the prior probability of fault is $\alpha$, then the probability of such partial assignment $\mathbf{s} \in \{0,1\}^n$ is $P(\mathbf{s}) = \alpha^{n_1}(1-\alpha)^{n-n_1}$, where $n_1$ is the number of faults in $\mathbf{s}$ (assuming that faults are independent). Test $T$ splits this effective state space into two sets, corresponding to outcome 0 (with probability mass $(1-\alpha)^n$) and outcome 1 (with probability mass $1-(1-\alpha)^n$) respectively. States within each set cannot be distinguished by $T$.

A natural alternative to selecting the most informative test is to pick the test that gives the most "balanced" partition of the current state space $\mathbf{S}^*$. Initially, when all states are indistinguishable, $\mathbf{S}^* = \{0,1\}^N$. Let $P(\mathbf{S}^* \mid T = 0)$ be the probability mass of states in $\mathbf{S}^*$ consistent with outcome 0 of test $T$. Similarly, define $P(\mathbf{S}^* \mid T = 1)$ for outcome 1. At every step, the next test is taken to be $\text{argmin}_T \left| P(\mathbf{S}^* \mid T = 0) - P(\mathbf{S}^* \mid T = 1) \right|$. After the outcome of $T$ becomes known, we discard the states in $\mathbf{S}^*$ inconsistent with this outcome.

The same greedy strategy was used by Dasgupta [3] in the context of actively learning a concept by adaptive queries, and by Kosaraju et al. [14] in a general setting. We show that this balance-based strategy is equivalent to the strategy based on choosing the most informative test. First we need to define the cost of a solution.

**Cost of diagnosis**: Both greedy strategies produce a tree with leaf nodes corresponding to possible states, and non-leaf nodes corresponding to tests, assuming that tests are informative enough to distinguish among the states (otherwise leaves correspond to distributions over subsets of states). The cost $c(\mathbf{s})$ of diagnosing leaf $\mathbf{s}$ in the set of leaves $\mathbf{S}$ is the number of tests on the path from the root to $\mathbf{s}$. The cost of a tree $D$ is given by $c(D) = \sum_{\mathbf{s}\in\mathbf{S}} P(\mathbf{s})c(\mathbf{s})$, the expected cost needed to diagnose a leaf chosen according to $P$.

**Equivalence**: Let $\mathbf{S}^*(\mathbf{t})$ be the set of states in $\mathbf{S}$ consistent with outcomes $\mathbf{t}$ of tests selected so far. For a test $T$, let $U$ be the set of leaves in $\mathbf{S}^*(\mathbf{t})$ consistent with outcome 0 of $T$. Also let $n_0 = |\overline{U}|$, $n_1 = |U|$, and $n = |\mathbf{S}^*(\mathbf{t})| = n_0 + n_1$.

To simplify the argument, assume that all weights in $\mathbf{S}^*(\mathbf{t})$ are equal; the argument readily holds for any set of weights.

The information gain is maximized by $T$ minimizing $\frac{1}{n}\sum_{i\in\{0,1\}} n_i \log n_i$. The most balanced split in this case is the one minimizing $|n_0 - n_1|$. Both functions are shown below. The V-shaped curve is the imbalance of a split with the corresponding $n_0$. The smooth flat curve is the conditional entropy of the state given the result of the split (and all previous splits). The remaining curve shows the ratio $\max\{n_0, n_1\}/\min\{n_0, n_1\}$, another equivalent splitting criteria. It is clear that the functions achieve their minimum at the same point.

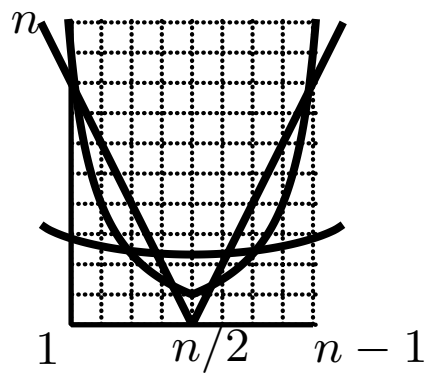

Both Dasgupta [3] and Kosaraju et al. [14] showed that the balance-based scheme results in a tree whose cost is within a factor of $\log(1/\min_{\mathbf{s}} P(\mathbf{s}))$ from optimal, where $\min_{\mathbf{s}} P(\mathbf{s})$ is the probability of the least probable state in $\mathbf{S}$. Furthermore, Kosaraju et al. [14] showed that a slight reweighting (needed if $P$ is exponentially unbalanced) results in a tree whose cost is within a factor of $O(\log N)$ from optimal, for any $P$. Thus a tree $D$ obtained by the greedy algorithm satisfies $c(D) \leq O(\log N)c(D^*)$, where $c(D^*)$ is the optimal cost.

As follows from the equivalence, the greedy algorithm that chooses the most informative split also results in a tree whose cost is within a factor of $O(\log N)$ from optimal.

Kosaraju et al. [14] also claimed that the guarantees hold for an algorithm that only *approximates* the most balanced partition. Assume without loss of generality that $P(\mathbf{S}^* \mid T = 0) \geq P(\mathbf{S}^* \mid T = 1)$, or in our case $n_0 \geq n_1$. The results says that if the best balance ratio $P(\mathbf{S}^* \mid T = 0)/P(\mathbf{S}^* \mid T = 1)$, or in our simple case, $n_0/n_1$, is approximated within a constant multiplicative factor, then the $O(\log N)$ approximation guarantee holds.

Let the most balanced split be $\{x^*, n - x^*\}$; without loss of generality, $x^* \geq n/2$. Assume that the approximate split is $\{x^* + a, n - x^* - a\}$. Notice that any approximation should be less balanced, thus $0 < a \leq n - 1 - x^*$. We want to upper bound the ratio:

$$\frac{(x^* + a)/(n - x^* - a)}{x^*/(n - x^*)} = 1 + \frac{an}{x^*(n - x^* - a)}.$$

Now $x^* \geq n/2$, and $n - x^* - a \geq 1$ since $a \leq n - 1 - x^*$. Thus the ratio is bounded by $1 + 2a$, which is constant if $a$ is constant.

It remains to show that good approximations of the conditional entropy result in small $a$'s. Note that *any* approximation of the conditional entropy is within a multiplicative factor of roughly $1 + (\log n)^{-1}$. Indeed, we have

$$\frac{x \log x + (n - x)\log(n - x)}{x^* \log x^* + (n - x^*)\log(n - x^*)},$$

which is maximized when $x$ is maximized and $x^*$ is minimized, or $x = n - 1$ and $x^* = n/2$, implying the upper bound.

The plot below shows the approximation ratio for the conditional entropy. Each curve corresponds to a particular exact split $x^*$ starting from $n/2$; here $n = 1000$. For every $x^*$, we plotted the approximation ratio as a function of approximate split $x = x^* + a$. Notice that the worst case ratio is roughly $1+1/\log(1000) \approx 1.1$, as expected. Also notice that if the approximate value is sufficiently close to the exact value, the curves are well approximated by lines (with larger exact values being harder to approximate). Thus for good approximations, the approximation ratio for conditional entropy translates roughly "linearly" into $a$. Thus, according to the result of Kosaraju et al. [14], we should expect a $O(\log N)$ approximation ratio even if entropies are only approximate.

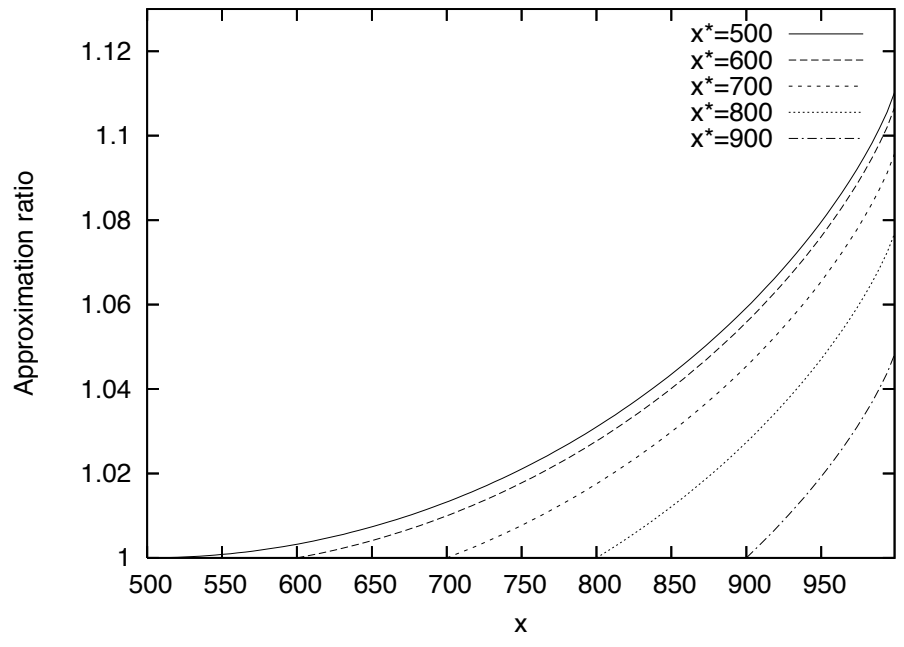

**State spaces**: Note that if all combinations of faults are possible, a simple information-theoretic argument shows that we need at least $N$ tests to uniquely distinguish between these states, which is as inefficient as testing each element directly. Of course, we can stop the diagnosis when the conditional entropy is sufficiently low (i.e., when we have an almost deterministic distribution on some subset of states), and then output the most likely state. This way we can often approximate the state with significantly fewer tests.

Another common approach is to assume an upper bound on the number of faults. For example, if we have prior fault probabilities $\alpha_i = P(S_i = 1)$, the expected number of faults is $\sum_{i=1}^{N} \alpha_i$ (assuming that faults are independent); hence if $\alpha_i = \alpha$ for all $i$, this number is $\alpha N$. By Markov's inequality, the probability that the actual number of faults is more than $\alpha N c$ is at most $1/c$ for any $c \geq 1$, thus we can typically assume that the state space $\mathbf{S}$ is the set of all subsets of at most $\alpha N c$ elements, for appropriate $c$.